\documentclass[10pt,journal,compsoc]{IEEEtran}
\ifCLASSOPTIONcompsoc
  \usepackage[nocompress]{cite}
\else
  \usepackage{cite}
\fi

\usepackage{times}
\usepackage{epsfig}
\usepackage{graphicx}
\usepackage{subfigure}
\usepackage{float}
\usepackage{amsmath}
\usepackage{amssymb}
\usepackage[ruled,linesnumbered]{algorithm2e} % For algorithms
\usepackage{algpseudocode}
\usepackage{multirow}
\usepackage{multicol}
\usepackage{array}
\usepackage{xcolor}
\usepackage{caption2}
\usepackage[normalem]{ulem}

\newcommand{\zjq}[1]{{\color{black}#1}}
\newcommand{\jqnew}[1]{{\color{black}#1}}

\newcommand{\shuyu}[1]{{\color{black}#1}}

\begin{document}
%
% paper title
% Titles are generally capitalized except for words such as a, an, and, as,
% at, but, by, for, in, nor, of, on, or, the, to and up, which are usually
% not capitalized unless they are the first or last word of the title.
% Linebreaks \\ can be used within to get better formatting as desired.
% Do not put math or special symbols in the title.
\title{Deep Line Art Video Colorization with a Few References}

\author{Min~Shi\thanks{\IEEEauthorrefmark{2} Both authors contributed equally to this research.}\IEEEauthorrefmark{2}, %~\IEEEmembership{Member,~ACM,}
	Jia-Qi~Zhang\IEEEauthorrefmark{2}, %~\IEEEmembership{Member,~ACM,}
    Shu-Yu~Chen, 
	Lin~Gao\thanks{\IEEEauthorrefmark{1} Corresponding Author.}\IEEEauthorrefmark{1}, %~\IEEEmembership{Member,~ACM,}
	Yu-Kun Lai, 
	Fang-Lue~Zhang
 	\IEEEcompsocitemizethanks{	
 	\IEEEcompsocthanksitem M. Shi and J. Zhang are with North China Electric Power University, Beijing 102206, China. E-mail:\{zhangjiaqi, shi\_ min\}@ncepu.edu.cn  
	\IEEEcompsocthanksitem S. Chen and L. Gao are ith the Beijing Key Laboratory of Mobile Computing and Pervasive Device, Institute of Computing Technology, Chinese Academy of Sciences, Beijing 100190, China, and also with the University of Chinese Academy of Sciences, Beijing 100190, China. E-mail:\{chenshuyu, gaolin\}@ict.ac.cn  
	\IEEEcompsocthanksitem Y.-K. Lai is with the School of Computer Science \& Informatics, Cardiff University, Wales, UK. Email: LaiY4@cardiff.ac.uk
	\IEEEcompsocthanksitem F. Zhang is with School of Engineering and Computer Science, Victoria University of Wellington, New Zealand. E-mail:fanglue.zhang@ecs.vuw.ac.nz 
 	}% <-this % stops an unwanted space
	%\thanks{Manuscript received August 10, 2018; revised August 26, 2018.}
}

% note the % following the last \IEEEmembership and also \thanks -
% these prevent an unwanted space from occurring between the last author name
% and the end of the author line. i.e., if you had this:
%
% \author{....lastname \thanks{...} \thanks{...} }
%                     ^------------^------------^----Do not want these spaces!
%
% a space would be appended to the last name and could cause every name on that
% line to be shifted left slightly. This is one of those "LaTeX things". For
% instance, "\textbf{A} \textbf{B}" will typeset as "A B" not "AB". To get
% "AB" then you have to do: "\textbf{A}\textbf{B}"
% \thanks is no different in this regard, so shield the last } of each \thanks
% that ends a line with a % and do not let a space in before the next \thanks.
% Spaces after \IEEEmembership other than the last one are OK (and needed) as
% you are supposed to have spaces between the names. For what it is worth,
% this is a minor point as most people would not even notice if the said evil
% space somehow managed to creep in.

%----------------
%----------------
% The paper headers
\markboth{}%
{Shell \MakeLowercase{\textit{et al.}}: Bare Demo of IEEEtran.cls for Computer Society Journals}
% The only time the second header will appear is for the odd numbered pages
% after the title page when using the twoside option.
%
% *** Note that you probably will NOT want to include the author's ***
% *** name in the headers of peer review papers.                   ***
% You can use \ifCLASSOPTIONpeerreview for conditional compilation here if
% you desire.

% The publisher's ID mark at the bottom of the page is less important with
% Computer Society journal papers as those publications place the marks
% outside of the main text columns and, therefore, unlike regular IEEE
% journals, the available text space is not reduced by their presence.
% If you want to put a publisher's ID mark on the page you can do it like
% this:
%\IEEEpubid{0000--0000/00\$00.00~\copyright~2015 IEEE}
% or like this to get the Computer Society new two part style.
%\IEEEpubid{\makebox[\columnwidth]{\hfill 0000--0000/00/\$00.00~\copyright~2015 IEEE}%
%\hspace{\columnsep}\makebox[\columnwidth]{Published by the IEEE Computer Society\hfill}}
% Remember, if you use this you must call \IEEEpubidadjcol in the second
% column for its text to clear the IEEEpubid mark (Computer Society jorunal
% papers don't need this extra clearance.)

% use for special paper notices
%\IEEEspecialpapernotice{(Invited Paper)}

% for Computer Society papers, we must declare the abstract and index terms
% PRIOR to the title within the \IEEEtitleabstractindextext IEEEtran
% command as these need to go into the title area created by \maketitle.
% As a general rule, do not put math, special symbols or citations
% in the abstract or keywords.
\IEEEtitleabstractindextext{%
\begin{abstract}
Coloring line art images based on the colors of reference images is an important stage in animation production, which is time-consuming and tedious. In this paper, we propose a deep architecture to automatically color line art videos with the same color style as the given reference images. Our framework consists of a \shuyu{color transform network and a temporal constraint network}. \shuyu{The color transform network} takes 
the target line art images as well as the line art and color images of one or more reference images as input, and generates corresponding target color images. To cope with larger differences between the target line art image and reference color images, our architecture utilizes non-local similarity matching to determine the region correspondences between the target image and the reference images, which are used to transform the local color information from the references to the target. To ensure global color style consistency, we further incorporate Adaptive Instance Normalization (AdaIN) with the transformation parameters obtained from a style embedding vector that describes the global color style of the references, extracted by an embedder. \shuyu{The temporal constraint network takes the reference images and the target image as its input in chronological order, and learns the spatiotemporal features through 3D convolution to ensure the temporal color consistency of the results.} Our model can achieve even better coloring results by fine-tuning the parameters with only a small amount of samples when dealing with an animation of a new style. To evaluate our method, we build a line art coloring dataset. Experiments show that our method achieves the best performance on line art video coloring compared to the state-of-the-art methods and other baselines. 
\end{abstract}

% Note that keywords are not normally used for peerreview papers.
}

% make the title area
\maketitle

% To allow for easy dual compilation without having to reenter the
% abstract/keywords data, the \IEEEtitleabstractindextext text will
% not be used in maketitle, but will appear (i.e., to be "transported")
% here as \IEEEdisplaynontitleabstractindextext when the compsoc
% or transmag modes are not selected <OR> if conference mode is selected
% - because all conference papers position the abstract like regular
% papers do.
\vspace{-0.7cm}
\IEEEdisplaynontitleabstractindextext
% \IEEEdisplaynontitleabstractindextext has no effect when using
% compsoc or transmag under a non-conference mode.

% For peer review papers, you can put extra information on the cover
% page as needed:
% \ifCLASSOPTIONpeerreview
% \begin{center} \bfseries EDICS Category: 3-BBND \end{center}
% \fi
%
% For peerreview papers, this IEEEtran command inserts a page break and
% creates the second title. It will be ignored for other modes.
\IEEEpeerreviewmaketitle

\IEEEraisesectionheading{\section{Introduction}\label{sec:introduction}}
% Computer Society journal (but not conference!) papers do something unusual
% with the very first section heading (almost always called "Introduction").
% They place it ABOVE the main text! IEEEtran.cls does not automatically do
% this for you, but you can achieve this effect with the provided
% \IEEEraisesectionheading{} command. Note the need to keep any \label that
% is to refer to the section immediately after \section in the above as
% \IEEEraisesectionheading puts \section within a raised box.

% The very first letter is a 2 line initial drop letter followed
% by the rest of the first word in caps (small caps for compsoc).
%
% form to use if the first word consists of a single letter:
% \IEEEPARstart{A}{demo} file is ....
%
% form to use if you need the single drop letter followed by
% normal text (unknown if ever used by the IEEE):
% \IEEEPARstart{A}{}demo file is ....
%
% Some journals put the first two words in caps:
% \IEEEPARstart{T}{his demo} file is ....
%
% Here we have the typical use of a "T" for an initial drop letter
% and "HIS" in caps to complete the first word.
\IEEEPARstart{T}{he} process of animation production requires high labor input. Coloring is one of the important stages after the line art images are created, which is a time-consuming and tedious task. Usually, ``inbetweeners'' colorize a series of line art images according to several reference color images drawn by artists. Some commercial devices and software can be used to speed up the workflow of line art image coloring. But it still needs a lot of repetitive work in each frame. Therefore, automatic methods for coloring line art images based on reference color images are highly demanded, which can greatly reduce the costs in animation production.

Early research on line art image coloring mostly relies on manually specifying colors, which are then spread out to similar regions~\cite{Qu:2006:MC:1141911.1142017,LazyBrush09,Levin:2004:CUO:1015706.1015780}. However, the efficiency of the coloring stage in animation production cannot be significantly improved using the above methods. Inspired by the success of generative models on image synthesis tasks in recent years, researchers have used deep convolutional neural networks (CNNs) to automatically color line art images ~\cite{Furusawa:2017:CSM:3145749.3149430,DBLP:journals/corr/ZhangJL17,DBLP:journals/corr/HensmanA17}. However, user interactions are required to achieve final satisfactory coloring results. To encourage temporal consistency in the colored animation, Thasarathan et al.~\cite{thasarathan2019automatic} input the previous frame together with the current frame to a discriminator to improve the colorization of neighboring frames. However, the model cannot fully guarantee the consistency between the color styles of their results and the reference image, which is important for the color quality of the animation.

Coloring line art videos based on given reference images is challenging. Firstly, unlike real life videos,  animation videos do not hold pixel-wise continuity between successive frames. For example, lines corresponding to limbs of an animated character may jump from one shape to another to depict fast motion.
As such temporal continuity cannot be directly used to guide the learning process to maintain the color consistency between regions in adjacent frames. Furthermore, line art images only consist of black and white lines, which lack rich texture and intensity information compared to grayscale images, making colorization a harder task. 
Also, in practice when coloring a new anime video from line art images, only a few examples are available. This requires the model to be trainable with a small number of samples, when applied to color animation videos with a new color style.

In this paper, we propose a deep architecture to automatically color line art videos based on reference images. In order to avoid possible error accumulation when continuously coloring a sequence of line art images, we determine the region correspondences between the target image and reference images by adopting a similarity-based color transform layer, which is able to match similar region features extracted by the convolutional encoders from the target and reference images, and use this to transform the local color information from the references to the target.
Then, to ensure the global color consistency, 
the Adaptive Instance Normalization (AdaIN)~\cite{huang2018multimodal} parameters are learned from the embeddings extracted for describing the global color style of the reference images. \jqnew{We further use a 3D convolutional network to refine the temporal color consistency between the coloring result and the reference images.}
It is the first time to utilize such a global and local network architecture to color line art images.  Moreover, our model can achieve even better coloring results by fine-tuning the parameters with only a small amount of samples when dealing with animations of a new color style.

\section{Related Work}

\subsection{Sketch colorization without references}

Early research on line drawing colorization ~\cite{Qu:2006:MC:1141911.1142017, LazyBrush09, Levin:2004:CUO:1015706.1015780} allows the user to specify color using brushes and then propagates the color to similar areas. The range of propagation can be determined by finding the region similarity or specified by the user. Qu et al. ~\cite{Qu:2006:MC:1141911.1142017} use a level-set method to find  similar regions, and propagate users' scribbles to those regions. Orzan et al.~\cite{orzan2008diffusion} require the users to set the gradient range of curves to control the range of their scribbles. Given a set of diffusion curves as constraints, the final image is generated by solving the Poisson equation. Those methods require a lot of manual interactions to achieve desired coloring results. With the development of deep learning, researchers have used it to achieve automatic or interactive coloring~\cite{ci2018user-guided, varga2017automatic}. However, a lot of manual interaction is still required to obtain reliable coloring results. 

\begin{figure}
    {\includegraphics[width=0.98\linewidth]{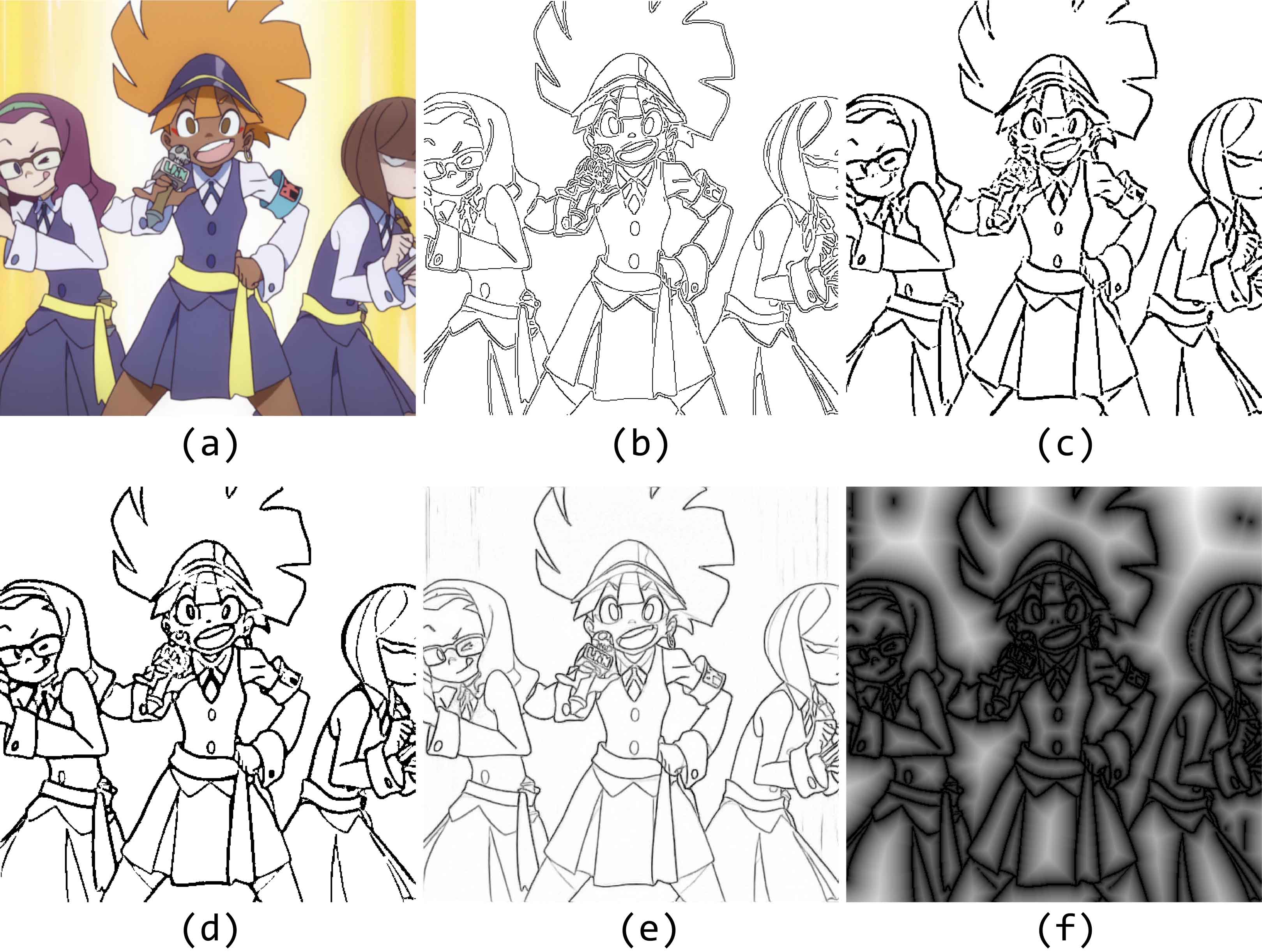}}
    \caption{Comparison of different line art image extraction methods. (a) original color image; (b) Canny ~\cite{canny1986a}; (c) XDoG ~\cite{winnemoller2012xdog}; (d) Coherent Line Drawing ~\cite{kang2007coherent}; (e) SketchKeras ~\cite{lllyasvielsketch}; (f) distance field map from SketchKeras results. }
    \label{fig:extract-line}
\end{figure}

\subsection{Sketch colorization with references}

To reduce manual workload, colorization methods with reference images have been proposed to achieve the specified color style for the target sketch. Sato et al.~\cite{sato2014reference-based} represent the relationships between the regions of line art images using a graph structure, and then solve the matching problem through quadratic programming. However, it is difficult to accurately segment complex line art images. Deep learning methods are proposed to 
avoid the requirement of accurate segmentation. Zhang et al. ~\cite{DBLP:journals/corr/ZhangJL17} extract VGG features from the reference image as their description, but their results have blurred object boundaries and mixed colors, probably due to the different characteristics of line art and normal images. To further refine the results, Zhang et al. ~\cite{zhang2019two-stage} divide the line art image colorization problem into the drafting stage and refinement stage, and users can input manual hints or provide a reference image to control the color style of the results ~\cite{lllyasvielstyle2paints}. However, these methods still require user interactions to achieve satisfactory coloring results. 
Hensman et al.~\cite{DBLP:journals/corr/HensmanA17} use cGANs (conditional Generative Adversarial Networks)
to colorize gray images with little need for user refinement, but the method is only applicable to learn relationships between grayscale and color images, rather than line art images. In addition, Liao et al. ~\cite{Liao:2017:VAT:3072959.3073683} use the PatchMatch~\cite{barnes2009patchmatch} algorithm to match the high-dimensional features extracted from the reference and target images, and realize style conversion between image pairs. It can be used for the line art image coloring task, but it does not match images with different global structures very well.

\begin{figure*}\centering

	{\includegraphics[width=0.98\linewidth]{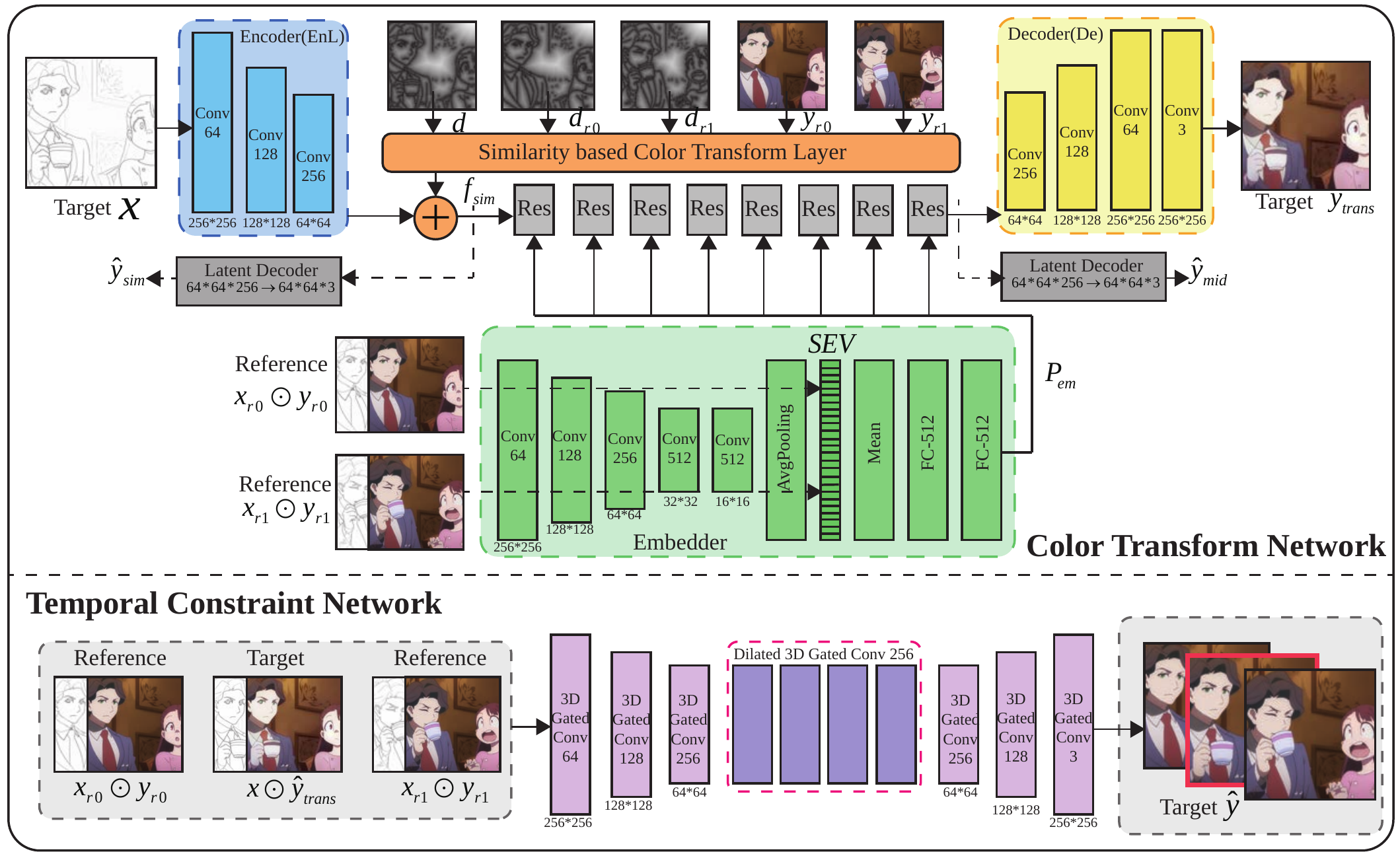}}
	\caption{The overall structure of the our network. \jqnew{The color transform network combines the latent features of global color style extracted from reference color images
	with latent features extracted from the input line art image to be colored. Finally, the temporal constraint network ensure the coloring result to be temporally consistent with the reference images.}
	}
	\label{fig:mian-network}
	\vspace{-5mm}
\end{figure*}

\subsection{Video colorization}

In video coloring research, animation coloring is much less explored than natural video coloring. Normal grayscale video coloring work~\cite{levin2004colorization, jampani2017video, zhang2019deep, Lei_2019_CVPR} learns temporal color consistency by calculating optical flow. However, animated video frames do not in general hold pixel-level continuity, causing optical flow algorithms to fail to get good results. Sýkora et al. ~\cite{Sykora:2004:UCB:987657.987677} propose a method which uses path-pasting to colorize black-and-white cartoons for continuous animation. However, when given sketch images contain lines which are not continuous, path-pasting is hard to find accurate correspondence. Automatic Temporally Coherent Video Colorization (TCVC)~\cite{thasarathan2019automatic} inputs the previous colored frame into the generator as a condition for the current line art image, and inputs both the previous and current colored frames into the discriminator to enforce temporal coherence. However, this method will cause error propagation in the generated coloring results, as errors introduced in one frame affect coloring results of all future frames. Instead, we use a method based on region matching with reference images to find the color features of the target image. Our method can better process long sequences of line art images without color error accumulation.

\section{Method}

In animation production, coloring in-between line art frames according to the key frames drawn and colored by the original artist is a necessary step. For automating this process, we design a generative network structure. 
\jqnew{One of our design goals is to maintain the local color consistency between the corresponding regions in the target and the reference images. We expect that our model will estimate accurate dense color transformations by matching local features between the target and reference line art images.} 
It involves a similarity-based color transform layer after the convolutional feature encoding stage. 
\jqnew{However, due to the possible large deformations of the same object in the line art images, it is not sufficient to obtain accurate coloring result by just matching local features of line art images. Therefore, we also expect the network to be capable of controlling the color style globally and correcting the coloring errors caused by inaccurate local similarity results. We thus propose to use a sub-module, Embedder, to extract the embeddings representing the global color style of the input reference images. The embedding vectors are used to ensure the global color style consistency between the generated coloring result and reference images. Finally, we add a 3D convolution network to improve the coloring quality by enhancing the temporal color consistency.}

\subsection{Data Preparation}\label{sec:dis-field}
Since there is no public dataset for the evaluation of  line art animation coloring, we built such a dataset to train our model and evaluate it. \zjq{We collect a set of animation videos and divide them into video shots using the color difference between the successive frames. The line art images corresponding to all the color frames of each video shot are extracted. See more details in section \ref{sec:DataCollection}.}

In order to obtain high-quality line art images from color animation, we tried various line extraction methods. The line art images extracted by the traditional Canny edge detection algorithm~\cite{canny1986a} are quite different from the line drawing images drawn by cartoonists. Other extractors, like XDoG edge extraction operator~\cite{winnemoller2012xdog} and the Coherent Line Drawing~\cite{kang2007coherent} method, are too sensitive to user-specified parameters. Sketch-Keras~\cite{lllyasvielsketch} is a deep line drafting model that uses a neural network trained to not only adapt to color images with different quality, but also to extract lines for the important details. Therefore, here we use the sketch-Keras to generate line art images, which have overall best lines, \jqnew{contain rich details and have the most similar style with that drawn by cartoonists.} Due to the data sparsity of lines in line art images, inspired by  SketchyGAN~\cite{chen2018sketchygan}, we also convert line art images into distance field maps to improve matching between images when training our model. The comparisons of different methods are shown in Fig.~\ref{fig:extract-line}.

\subsection{Overall Network Architecture}

\begin{figure*}\centering
	{\includegraphics[width=0.98\linewidth]{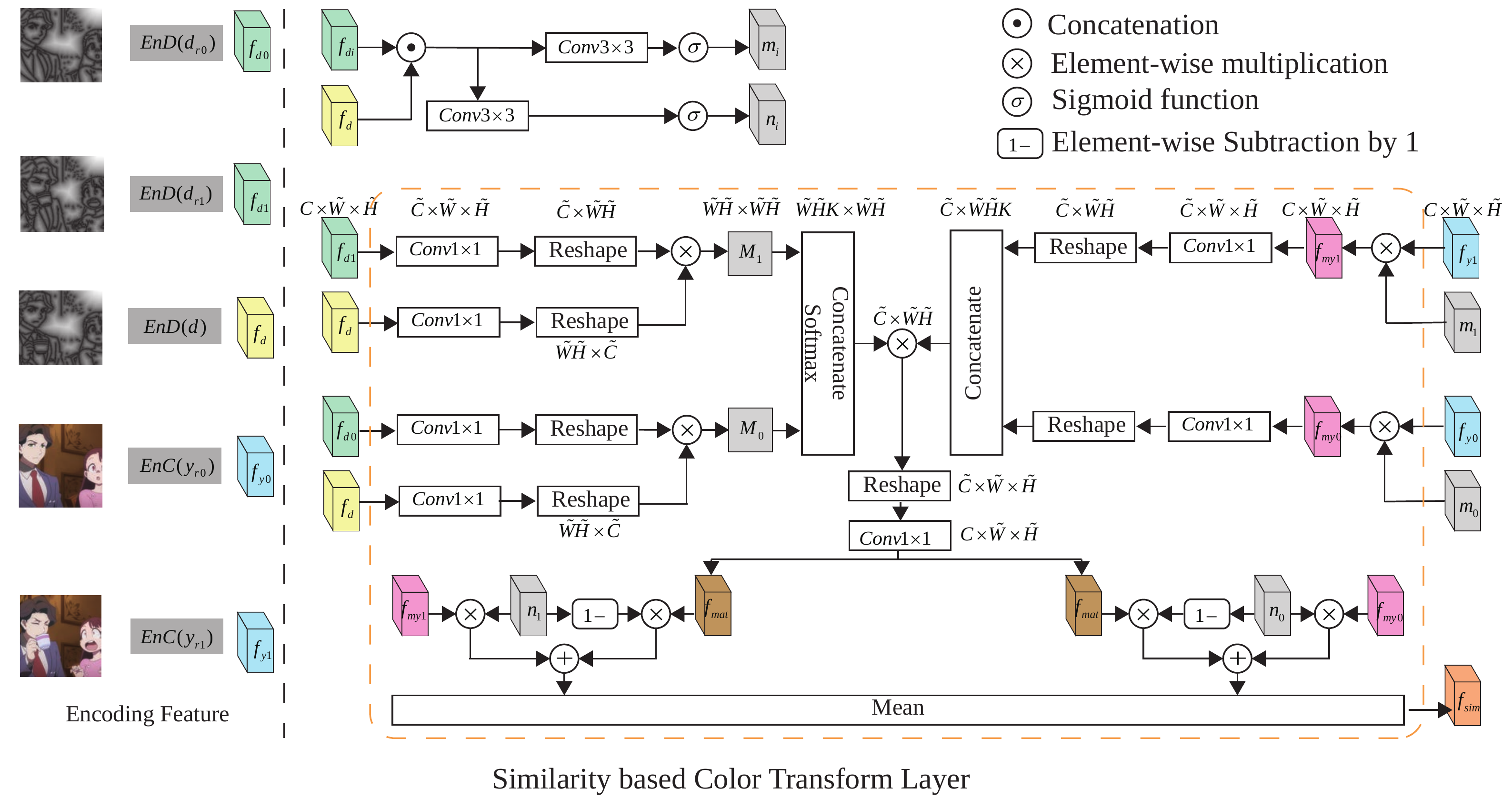}}
	\caption{Similarity-based color transform layer. It first extracts the feature maps of the input distance field maps of the target and reference images through the encoder $EnD$. Then the similarity map is calculated, which is for transforming the color features of the reference images to those of the target image $f_{mat}$. \jqnew{Finally, the matching-based transform feature $f_{mat}$ and the color feature of the reference images $f_{myi}$ are dynamically combined to obtain the local color feature map of the target image $f_{sim}$.}
	}
	\label{fig:similar-match}
	\vspace{-5mm}
\end{figure*}

\jqnew{Our network consists of two sub-networks, color transform network and temporal constraint network. The color transform network} consists of a conditional generator $G$ and an adversarial discriminator $D$. Inspired by the image-to-image translation framework~\cite{liu2019few-shot}, \jqnew{our generator learns global style information through Embedder, and provides local color features through the similarity-based color transform layer.} Thus, our generator $G$ takes a target line art image $x$, a target distance field map $d$ and two reference images (including the line art, distance field map and color image for each reference image) $\{x_{r_0},x_{r_1};d_{r_0},d_{r_1};y_{r_0},y_{r_0}\}$ as input and produces the target color image $\hat{y}_{trans}$ via
\begin{equation}
\hat{y}_{trans}  = G(x,d,\{x_{r_0},x_{r_1};d_{r_0},d_{r_1};y_{r_0},y_{r_1}\})
\label{equ:target}
\end{equation}
\jqnew{$r$ represents the reference object. $\hat{y}_{trans}$ is the preliminary coloring result generated by the color transform network, where "0" and "1" represent the beginning and end of the video sequence respectively.}

Figure~\ref{fig:mian-network} shows the architecture of the color transform network, which is mainly composed of six parts,  encoders, a similarity-based color transform layer $Sim$, a group of middle residual blocks $Mid$, a style information Embedder $Em$, a decoder $De$, and a discriminator $D$. First, we use the encoders to extract the feature maps of the following images: the target line art image $EnL(x)$, the target distance field map $EnD(d)$, the reference distance field maps $EnD(d_{r_0},d_{r_1})$, and the reference color images $EnC(y_{r_0},y_{r_1})$. $EnL$, $EnD$, and $EnC$ are three identically constructed encoders that extract the features of line art images, field distance maps, and color images, respectively. The extracted feature maps of the above distance field maps and color images are then fed to the similarity-based color transform layer to obtain the local color features $f_{sim}$ of the target line art image, which provide a local colorization for the target image. After concatenating the line art images and the reference color images, we also feed them separately into an Embedder module~\cite{liu2019few-shot} to get intermediate latent vectors for reference images, and then compute the mean of their intermediate latent vectors to obtain the final style embedding vector (SEV) $Em(x_{r_1} \odot y_{r_1},...,x_{r_K} \odot y_{r_K})$. The SEV is used to adaptively compute the affine transformation parameters $P_{em}$ for the adaptive instance normalization (AdaIN) residual blocks~\cite{huang2018multimodal} via a two-layer fully connected network, which learns the global color information for the target image. Then, with the SEV controlling the AdaIN parameter, the output features of the $Sim$ layer 
(capturing similarity matching between the target and references)
and $EnL$ (capturing local image information of the target) are added to get the input to the $Mid$ module, whose output then forms the input to the decoder $De$ to generate the final target color image $\hat{y}_{trans}$. Eq.~\ref{equ:target} is factorized to
\begin{align}
\nonumber    f_{sim} =& Sim(EnD(d),\{EnD(d_{r0}),EnD(d_{r1}); \notag \\
\nonumber &       EnC(y_{r0}),EnC(y_{r1})\}), \notag \\
\nonumber    P_{em} =& Em(x_{r0} \odot y_{r0},x_{r1} \odot y_{r1}), \notag
    \\
    \hat{y}_{trans} =& De(Mid(EnL(x) \oplus f_{sim}, P_{em})),
\end{align}
where $\odot$ means concatenation, and $\oplus$ means element-wise summation.
The design of our color transform network ensures that the coloring process considers both the local similarity with the reference image regions and the global color style of the reference images.

The color transform network discriminator $D$ takes as input an image pair of line art image $x$ and corresponding color image, either real color image $y$ from the training set, or $\hat{y}_{trans}$ produced by the generator $G$. It is trained to solve a binary classification task to determine whether the color image is real or fake (generated). The discriminator is trained in an adversarial way to try to make the generated images indistinguishable from real color animation images.

\jqnew{
%To improve the continuity of the coloring results of the line art video sequence while avoiding error accumulation, 
To make the coloring result temporally coherent, we apply a 3D convolutional generation adversarial network to learn the temporal relationship between the target coloring image and the reference images. The generator takes the reference and target line art and color image pairs as its input, and generates the target coloring result $\hat{y}$ as well as the reference beginning and the end frames. The input and output are both put in chronological order. The discriminator is trained to perform a binary classification to determine whether the color image sequence are real or fake. In order to reduce the training time and parameter amount of 3D convolution, we use the learnable gated temporal shift module\cite{chang2019free} (LGTSM) based on temporal shift module (TSM) in both of the generator and discriminator. The LGTSM structure uses 2D convolution to achieve the effect of 3D convolution with guaranteed performance. }

\subsection{Similarity-based Color Transform Layer}\label{sec:similar}

To cope with larger changes between adjacent frames in sketches,
existing work~\cite{thasarathan2019automatic} learns the matching of adjacent frames by increasing the number of intermediate residual blocks in the generator network. However, our method needs to match references with target line art drawing, which can have even more significant differences.
\jqnew{ We therefore utilize the global relationship learned by non-local neural networks~\cite{wang2018non-local} used in the work of video colorization~\cite{zhang2019deep}.
Compared with matching features from grayscale images ~\cite{zhang2019deep}, matching features from line art images does not directly obtain satisfactory results because of the large deformations of lines of objects. We expect the network to learn the confidence of the correspondences simultaneously. Therefore, we make the similarity-based color transform layer to additionally learn masks of the features of reference images, which adaptively select the positions in the reference color images with the highest matching confidence. In the last stage of this module, we dynamically combine the matching-based transform feature from the whole reference images and the color feature, where the color features are selected by the masks based on the matching confidence, and the matching-based transform feature provides features learned from the whole reference images.}

\jqnew{The overall structure of the similarity-based color transform layer is shown in Fig.~\ref{fig:similar-match}. We calculate the similarity of the high-dimensional features extracted from the target line art image and the reference line art images on a global scale. The module uses two learned internal masks $m_i$ and $n_i$:
\begin{equation}
m_{i} = \sigma(Conv(f_{d_i} \odot f_d))
\end{equation}
\begin{equation}
n_{i} = \sigma(Conv(f_{d_i} \odot f_d))
\end{equation}
where $f_d$ is the features from the target, and $f_{d_i}$ is the $i$-th reference image ($i=0,1$). The mask $m_i$ is used to select new features $f_{my_i}=f_{y_i} \otimes m_i$ from the reference color image features $f_{y_i}$, and $n_i$ is used to combine the features matching information $f_{mat}$ with the reference color image features $f_{my_i}$. 

The matching-based transform feature $f_{mat}$ is obtained through estimating the similarity between the features of target and reference line art images. To reduce the complexity of global feature matching, we apply $1 \times 1$ convolutions to reduce the number of channels for feature maps to $\tilde{C}=C/8$, and reshape them to size $\tilde{W}\tilde{H} \times \tilde{C}$. The similarity map $M_k$ ($\tilde{W}\tilde{H} \times \tilde{W}\tilde{H}$) measures the similarity between features at different locations of the feature maps of the target and the $i$-th reference image. It is obtained through matrix multiplication:
$M_i = f_d \cdot f_{d_i}^T$. We concatenate the matrices for all the reference images and apply softmax to $\{M_i\}$ to form the matching matrix $\tilde{M}$ of size $\tilde{W}\tilde{H}K \times \tilde{W}\tilde{H}$. Similarly, we apply $1\times 1$ convolutions to reduce the channels of the color feature map of the reference images, reshape and concatenate them to form reference color matrix $f_C$ of size $\tilde{C} \times \tilde{W}\tilde{H}K$. The output of the module, $f_{mat} = f_C \cdot \tilde{M}$, represents the matching-based transform feature which transforms the color information from the reference images to the target based on the local similarity.

We use the following approach to get the final color similarity feature $f_{sim}$. :
\begin{equation}
f_{sim} = \frac{1}{2}\sum_{i=0}^{1}( (1-n_i)\otimes f_{mat} + f_{my_i} \otimes n_i )
\label{equ:sim_feature}
\end{equation}

To make matching more effective, all the input line art images are first turned into distance field maps, which are used for extracting feature maps using the encoder $EnD$. Similarly, the color information of reference images $y_{r_0}$ and $y_{r_1}$ are extracted through encoders $EnC$. Let $W$ and $H$ be the width and height of the input images, and $\tilde{W}=W/4$, $\tilde{H}=H/4$, and $C$ are the width, height and channel number of the feature maps. 
The feature map size is reduced to $1/4$ of the input image size to make a reasonable computation resource demand of similarity calculation.
}

\begin{figure*}\centering

    {\includegraphics[width=0.98\linewidth]{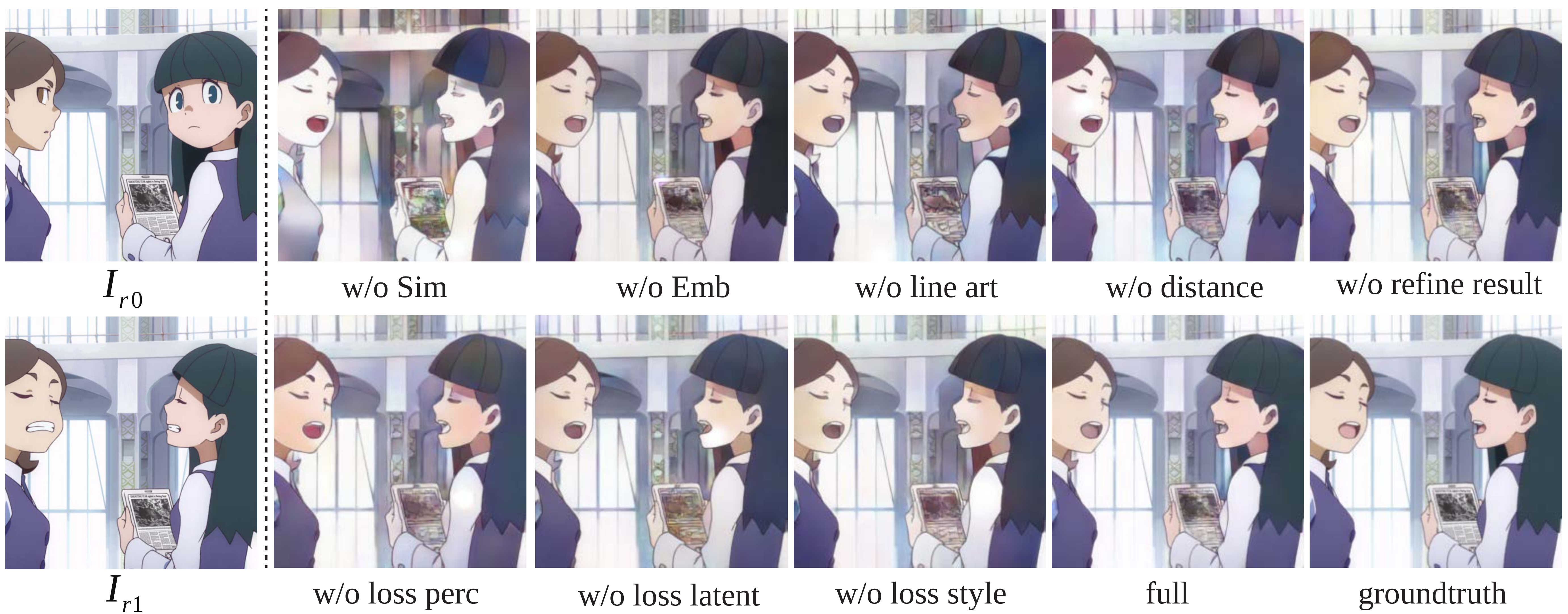}}
    \vspace{-3mm}
    \caption{Network model component analysis.}
	\label{fig:ana-component}
	\vspace{-3mm}
\end{figure*}

\begin{figure*}[t]
\centering
% \vspace{-2mm}
    \subfigure[]{\includegraphics[width=0.13\linewidth]{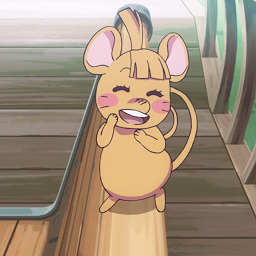}}
    \subfigure[]{\includegraphics[width=0.13\linewidth]{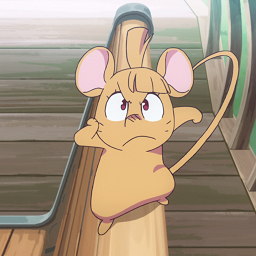}}
    \subfigure[]{\includegraphics[width=0.13\linewidth]{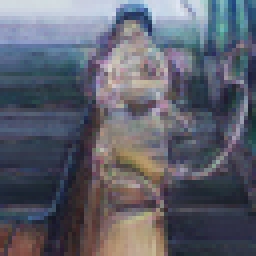}}
    \subfigure[]{\includegraphics[width=0.13\linewidth]{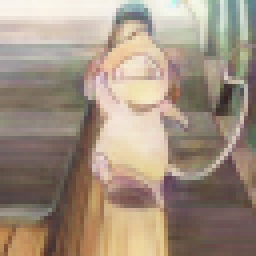}}
    \subfigure[]{\includegraphics[width=0.13\linewidth]{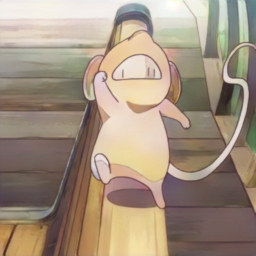}}
    \subfigure[]{\includegraphics[width=0.13\linewidth]{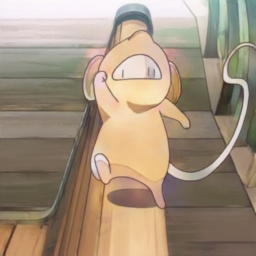}}
    \subfigure[]{\includegraphics[width=0.13\linewidth]{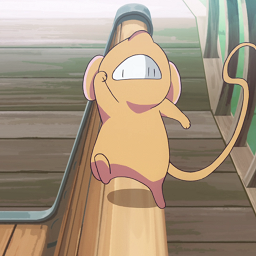}}\\
% 	\vspace{-0.4cm}
\vspace{-2mm}
	\caption{Intermediate results. (a) Reference color image 1, (b) Reference color image 2, (c) Latent decoder image $\hat{y}_{Sim}$, (d) Latent decoder image $\hat{y}_{Mid}$, (e) No temporal color image, (f) Final color image, (g) Ground truth.}
	\label{fig:color-decoder}
% 	\vspace{-6mm}
\end{figure*}

\begin{figure}[t]
\centering
% \vspace{-2mm}
    
    {\includegraphics[width=0.3\linewidth]{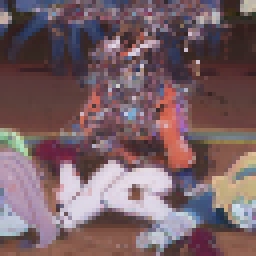}}
    {\includegraphics[width=0.3\linewidth]{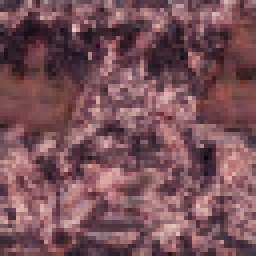}}
    {\includegraphics[width=0.3\linewidth]{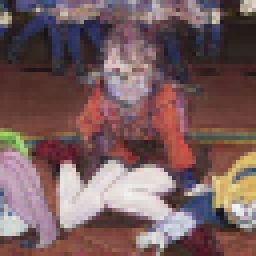}}\\
    
    \subfigure[]{\includegraphics[width=0.3\linewidth]{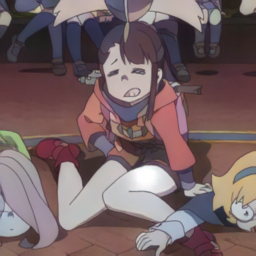}}
    \subfigure[]{\includegraphics[width=0.3\linewidth]{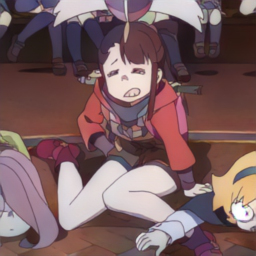}}
    \subfigure[]{\includegraphics[width=0.3\linewidth]{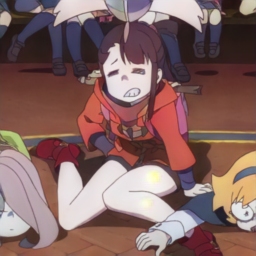}}\\
    
	\caption{Sim module reconstruction results. The first raw represents the network generation results, and the second raw represents the Sim module reconstruction results. (a) All convolutional layers use spectral normalization; (b) All convolutional layers do not use spectral normalization; (c) Only Encoders and Decoder use spectral normalization.}
	\label{fig:color-sim}
% 	\vspace{-6mm}
\end{figure}

% UNet; cGAN; TCVC;

\begin{figure*}\centering
	{\includegraphics[width=0.12\linewidth]{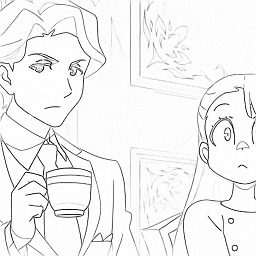}}
	{\includegraphics[width=0.12\linewidth]{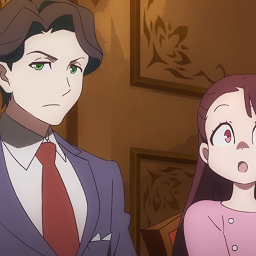}}
    {\includegraphics[width=0.12\linewidth]{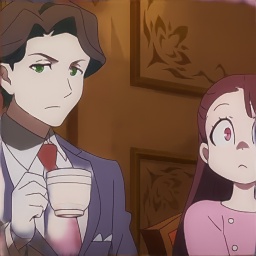}}
    {\includegraphics[width=0.12\linewidth]{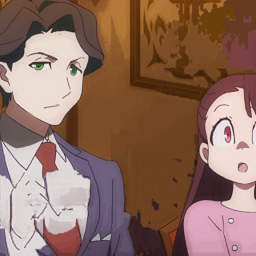}}
    {\includegraphics[width=0.12\linewidth]{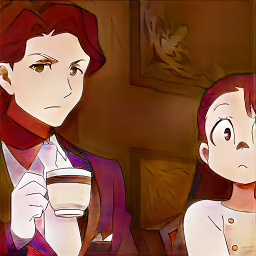}}
    {\includegraphics[width=0.12\linewidth]{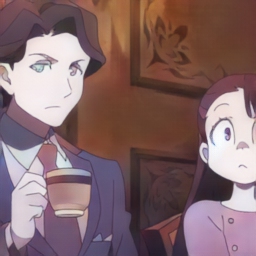}}
	{\includegraphics[width=0.12\linewidth]{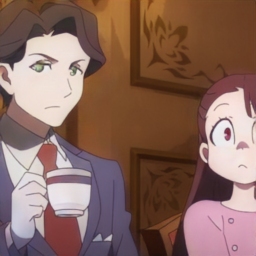}}
    {\includegraphics[width=0.12\linewidth]{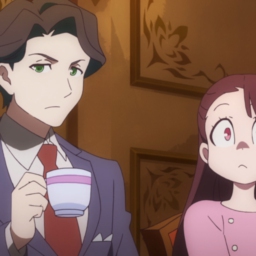}}
	
	{\includegraphics[width=0.12\linewidth]{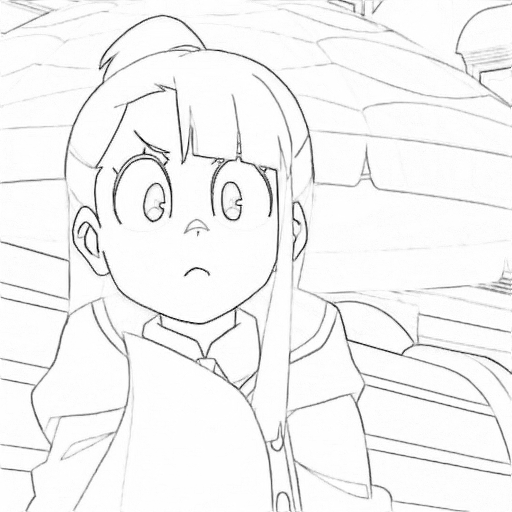}}
	{\includegraphics[width=0.12\linewidth]{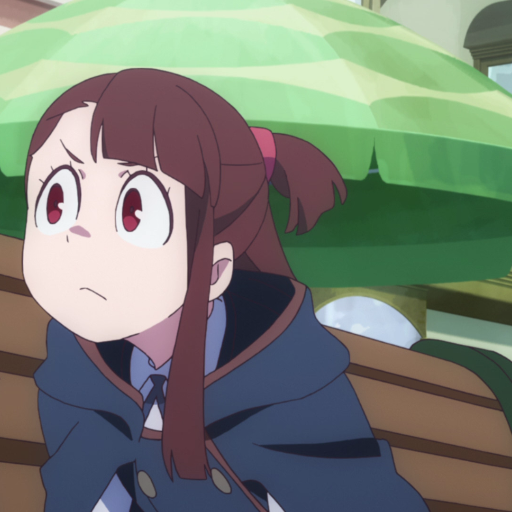}}
    {\includegraphics[width=0.12\linewidth]{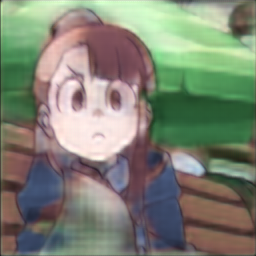}}
	{\includegraphics[width=0.12\linewidth]{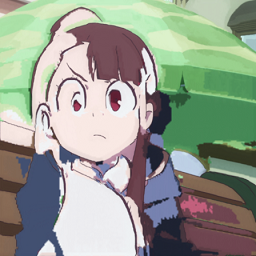}}
    {\includegraphics[width=0.12\linewidth]{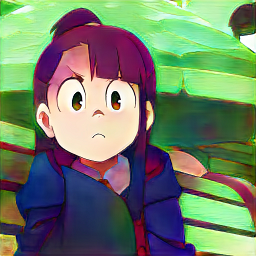}}
    {\includegraphics[width=0.12\linewidth]{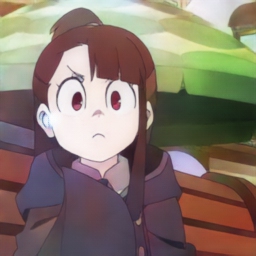}}
	{\includegraphics[width=0.12\linewidth]{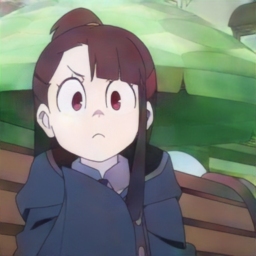}}
    {\includegraphics[width=0.12\linewidth]{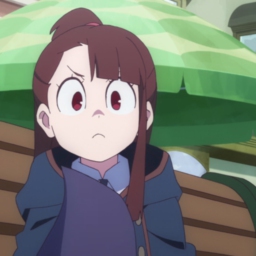}}\\

	\vspace{-0.5em}
	\subfigure[]{\includegraphics[width=0.12\linewidth]{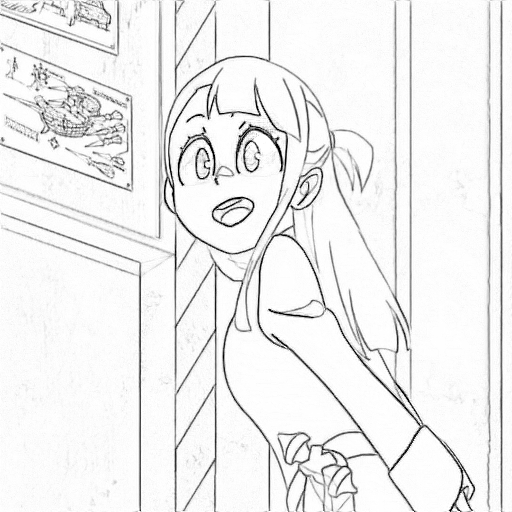}}
	\subfigure[]{\includegraphics[width=0.12\linewidth]{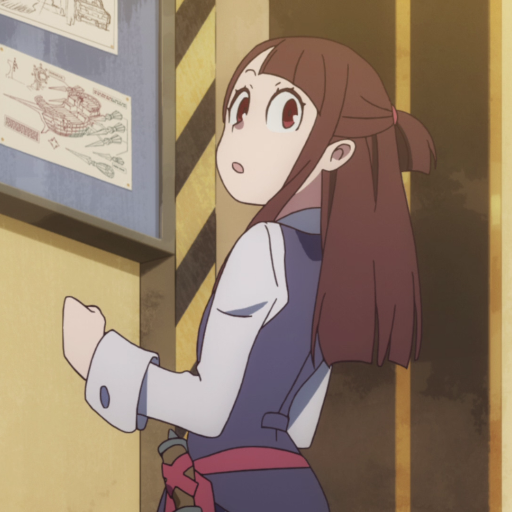}}
    \subfigure[]{\includegraphics[width=0.12\linewidth]{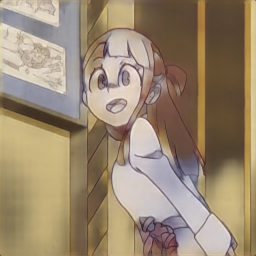}}
	\subfigure[]{\includegraphics[width=0.12\linewidth]{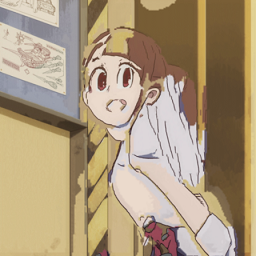}}
    \subfigure[]{\includegraphics[width=0.12\linewidth]{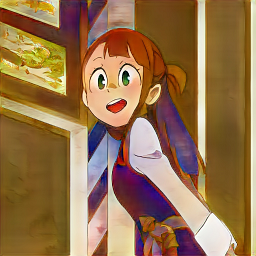}}
    \subfigure[]{\includegraphics[width=0.12\linewidth]{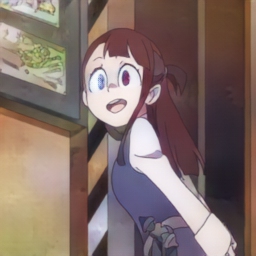}}
	\subfigure[]{\includegraphics[width=0.12\linewidth]{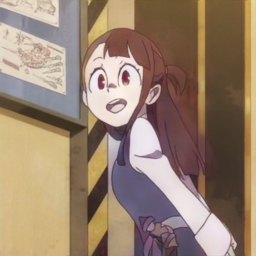}}
    \subfigure[]{\includegraphics[width=0.12\linewidth]{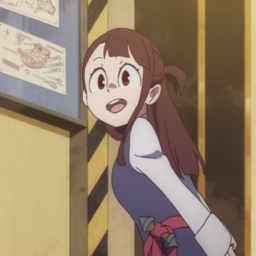}}

    \caption{Comparison with existing methods. (a) input sketch image; (b) the reference image;  (c) results of cGAN~\cite{DBLP:journals/corr/HensmanA17}; (d) results of Deep Image Analogy~\cite{Liao:2017:VAT:3072959.3073683}; (e) results of Two-stage method~\cite{zhang2019two-stage}; (f) results of TCVC~\cite{thasarathan2019automatic}; (g) our results; (h) Ground truth. }
	\label{fig:ana-other}
% 	\vspace{-5mm}
\end{figure*}

\subsection{Loss}
To enforce the color coherence between similar regions and penalize the color style difference with the reference images, we define our loss function based on the following loss terms. \jqnew{Meanwhile, in the temporal constraint network, the generation result of the network is divided into multiple images and the following loss terms is applied. Finally, the calculated loss values of multiple images are averaged to obtain the final loss value.}

\textbf{L1 loss}. To encourage the generated results to be similar to the ground truth, we use the pixel level L1 loss measuring the difference between the network generated result $\hat{y}$ and the ground truth color image $y$:
\begin{equation}
\mathcal{L}_{L1}  = \Arrowvert y - \hat{y} \Arrowvert_{1}
\end{equation}

\textbf{Perceptual loss}. In order to ensure that the generated results are perceptually consistent with ground truth images, we use the perceptual loss introduced in 
~\cite{johnson2016perceptual}:
%the real images on the high-dimensional features:
\begin{equation}
\mathcal{L}_{perc}  = \sum_{i=1}^{5}\frac{1}{N_i}\Arrowvert \Phi_{\hat{y}}^{i} - \Phi_{y}^{i} \Arrowvert_2^2
\end{equation}
where  $\Phi^{i}$ ($i=1,\dots,5)$ represents the feature map extracted at the ReLU $i$\_1 layer from the VGG19 network~\cite{simonyan2014very}. For our work, we extract feature maps from the VGG19 network layers ReLU 1\_1; ReLU2\_1; ReLU 3\_1; ReLU 4\_1 and ReLU 5\_1. $N_i$ represents the number of elements in the $i$-th layer feature map.

\textbf{Style loss}. Similar to image style conversion work~\cite{gatys2016image, risser2017stable, thasarathan2019automatic}, we calculate the similarity of the Gram matrices on the high-dimensional feature maps of the generated image and the ground truth to encourage the generated result to have the same style as the ground truth.
\begin{equation}
\mathcal{L}_{style}  = \sum_{i=1}^{5}\Arrowvert G(\Phi_{\hat{y}}^{i}) - G(\Phi_{y}^{i}) \Arrowvert_{1},
\end{equation}
where $G(\cdot)$ calculates the Gram matrix for the input feature maps. We use the VGG19 network to extract image feature maps from the same layers for calculating perceptual loss.

\textbf{Latent constraint loss}. In order to improve the stability of the generated effect, inspired by~\cite{DBLP:journals/corr/ZhangJL17, kim2019tag2pix}, in addition to constraining the final generation results, we introduce further  constraints on intermediate results of the network. Specifically, we add multi-supervised constraints to the similarity-based color transform layer output $f_{sim}$ and $Mid$ module output $f_{mid}$.

\jqnew{
To make perceptual similarity measured more easily, $f_{sim}$ and $f_{mid}$ first pass through latent decoders (involving a convolution layer) to output 3-channel color
images $\hat{y}_{sim}$ and $\hat{y}_{mid}$. We then use L1 loss to measure their similarity with the ground truth as follows:

\begin{equation}
\mathcal{L}_{latent}  = \Arrowvert y - \hat{y}_{sim} \Arrowvert_{1} + \Arrowvert y - \hat{y}_{mid} \Arrowvert_{1}
\end{equation}

}

% Adversarial Loss.

\textbf{Adversarial Loss}. The adversarial loss promotes correct classification of real images ($y$) and generated images ($\hat{y}$).
% \begin{equation}
\begin{gather}
    \mathcal{L}_{GAN}(G, D) = \mathbb{E}_{(x, y)}[\log D(x, y)] +  \notag \\
        \mathbb{E}_{(x, \hat{y})}[\log(1 - D(x, \hat{y}))]
\end{gather}
% \end{equation}

% Objective Function.
\textbf{Overall objective loss function}. Combining all of the above loss terms together, we set the optimization goal for our model:
\begin{gather}
\mathcal{L}_{x} = \lambda_{perc}\mathcal{L}_{perc} + \lambda_{style}\mathcal{L}_{style} + \notag \\ 
\lambda_{latent}\mathcal{L}_{latent} + \lambda_{GAN}\mathcal{L}_{GAN} + \lambda_{L1}\mathcal{L}_{L1}
\end{gather}
where $\lambda$ controls the importance of terms. We set $\lambda_{perc} = 1$, $\lambda_{style} = 1000$, $\lambda_{latent} = 1$, $\lambda_{GAN} = 1$, $\lambda_{L1} = 10$. \zjq{Since the resulted style loss value is relatively small in our experiments, we set its weight as $1000$, to make its contribution comparable with the GAN loss.}

\subsection{Implementation details}

\jqnew{Our network consists of two sub-networks, color transform network and temporal constraint network. We first train the color transform network to ensure that the network generates plausible coloring results. Then the color transform network parameters are fixed, and the temporal constraint network parameters are optimized to refine the temporal consistency between the coloring result and the reference images.}

The color transform network is composed of a generator network and a discriminator network. The generator network consists of encoders, a similarity-based color transform layer, middle residual convolution blocks, a Decoder, and an Embedder. The encoders for line art images, distance field maps and color images share the same network architecture. 
They are composed of $3$ convolution layers, and use instance normalization since colorization should not be affected by the samples in the same batch. They utilize the ReLU activations. We have $8$ middle residual blocks~\cite{huang2018multimodal} with AdaIN ~\cite{huang2017arbitrary} as the normalization layer and ReLU activations. The Decoder is composed of $4$ convolutional layers, also with the instance normalization and the ReLU activations.
Before the convolution of Decoder, we use the nearest neighbor upsampling to enlarge the feature map by $2$ times along each spatial dimension. This will eliminate the artifacts of checkerboard pattern in the generated results of GAN~\cite{odena2016deconvolution}. The Embedder consists of $5$ convolution layers followed by a mean operation along the sample axis. Specifically, it maps multiple reference images to latent vectors and then averages them to get the final style latent vector. The affine transformation parameters are adaptively computed using the style latent vector by a two-layer fully connected network. 
\jqnew{Meanwhile, encoders and decoder apply spectral normalization \cite{DBLP:conf/iclr/MiyatoKKY18} to their convolutional layers to increase the stability of training.} 
The discriminator network structure is similar with Embedder, which consists of $5$ layers of convolutions. At the same time, in addition to the last layer of convolution, the discriminator adds spectral normalization \cite{DBLP:conf/iclr/MiyatoKKY18} to other convolutional layers. It utilizes the Leaky LeakyReLU activations.

\jqnew{The temporal constraint network is composed of a generator and a patch discriminator. The generator is composed of an Encoder with $3$ 3D gated convolutional layers, and a Decoder with $4$ dilatd 3D gated convolutional layers and $3$ 3D gated convolutional layers. In the last $3$ convolutional layers of the Decoder, we use nearest neighbor upsampling to enlarge feature maps by $2$ times along each spatial dimension. The patch discriminator is composed of $5$ 3D convolutional layers. The spectral normalization is applied to both the generator and discriminator to enhance training stability.}

\begin{figure}\centering
	{\includegraphics[width=0.98\linewidth]{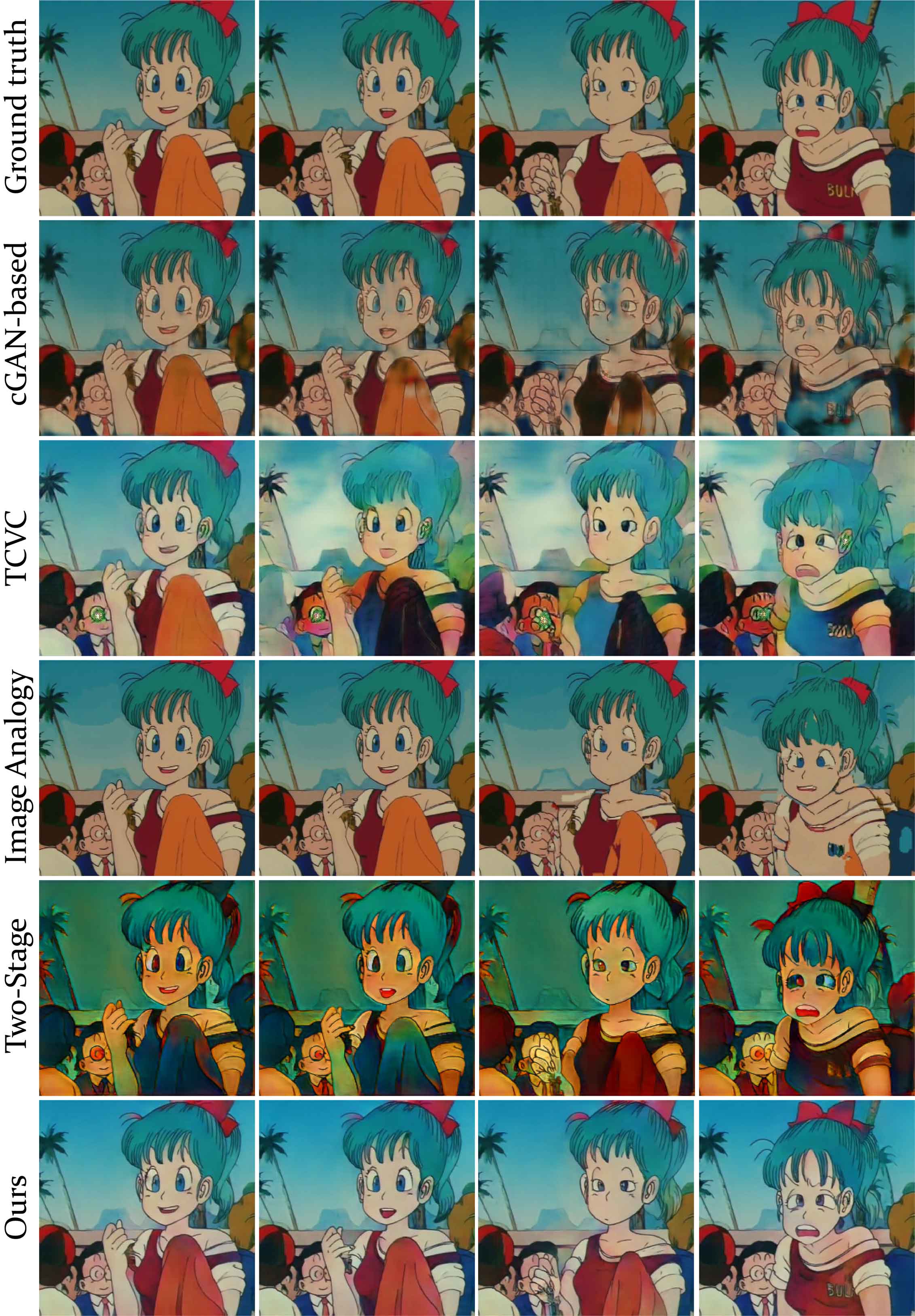}}
	 \caption{Video sequence coloring result comparison. We compare the method with cGan-based ~\cite{DBLP:journals/corr/HensmanA17}, Deep Image Analogy ~\cite{Liao:2017:VAT:3072959.3073683}, Two-stage ~\cite{zhang2019two-stage}, TCVC ~\cite{thasarathan2019automatic} on a long sequence of coloring results. We show several frames from the long sequence at an equal interval.
	}
	 \label{fig:ana-ball}
% 	\vspace{-3mm}
\end{figure}

\section{Experiments}

We evaluate the line art video coloring results on our line art colorization dataset. We show that our model outperforms other methods both quantitatively and qualitatively. In the following, we first introduce the data collection details. Then we analyze the effectiveness of the various components of the proposed method, and report the comparison between our method and the state-of-the-art methods both qualitatively and quantitatively.

\begin{figure}\centering
	{\includegraphics[width=0.98\linewidth]{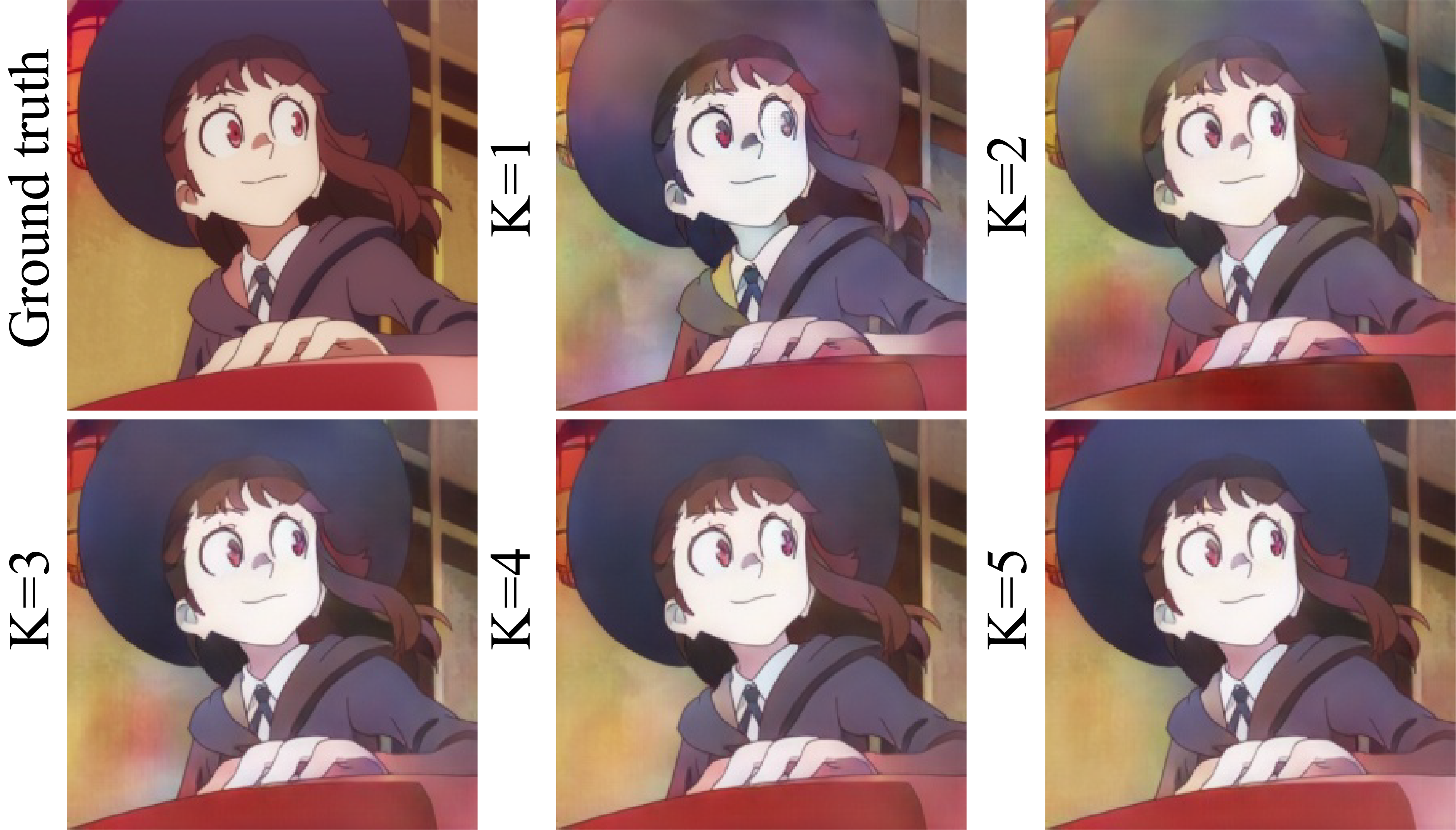}}
% 	\vspace{-1mm}
    \caption{Comparison of the coloring results with different numbers of reference images $K$.} 
	\label{fig:ana-diffseq}
% 	\vspace{-5mm}
\end{figure}

\subsection{Data Collection and Training}\label{sec:DataCollection}

We extract video frames from selected animations and extract the line art images to form our training dataset. We calculate a 768-dimensional feature vector of histograms of R, G, B channels for each frame. The difference between frames is determined by calculating the mean square error of the feature vectors, which is used for splitting the source animations into shots. When the difference between the neighboring frames is greater than 200, it is considered to belong to different shots. In order to improve the quality of the data, we remove shots in which the mean square errors between all pairs of frames are less than 10 (as they are too uniform), and the shot with a length less than 8 frames. Then we filter out video frames that are too dark or too faded in color. Finally we get a total of 1096 video sequences from 6 animations, with a total of 29,834 images. \zjq{Each video sequence has 27 frames on average.} These 6 anime videos include: 1) Little Witch Academia; 2) Dragon Ball; 3)  Amanchu; 4) SSSS.Gridman; 5) No.6; 6) Soul Eater. All the images are scaled to   $256\times256$ size. 

To evaluate the versatility, we choose the data of Anime 1, a total of 416 video sequences, 11,241 images for training and testing the network. Specifically, 50\% of the data in the Anime 1 is used as the training data, and the remaining 50\% is used as the test data. 
Other anime data is mainly used for testing, apart from using a few sequences for fine-tuning. 

We use the Adam optimizer and set the generator learning rate to $1\times 10^{-4}$, the discriminator learning rate to $1\times10^{-5}$, $\beta_{1}$ to 0.5, $\beta_{2}$ to 0.999, batch size 4. The color transform network trained 40 epochs, temporal constraint network trained 10 epochs. The experiment is performed on a computer with an Intel i7-6900K CPU and a GTX 1080Ti GPU, and the training time is about 2 days. \zjq{It takes 71ms to color a line art image.} 

\jqnew{The number of frames in each video sequence could vary from 8. Thus we randomly extract 8 successive frames from videos for network training, where the first and last frames are used as reference images, and the intermediate frames are used as the target images. When testing other methods, the first frame of the entire sequence is selected as the reference image to color the other frames.}

\begin{table}
\begin{center}
\begin{tabular}{|l|c|c|c|c|}
\hline
Setting & MSE$\downarrow$ & PSNR$\uparrow$ & SSIM$\uparrow$ & FID$\downarrow$ \\
\hline
 No Sim &  0.0220 &  17.27 &  0.77 &  48.37 \\
 No Emb &  0.0149 &  19.92 &  0.86 &  26.97 \\
 No Line Art &  0.0198 &  18.71 &  0.77 &  37.06 \\
 No Distance &  0.0126 &  20.46 &  0.84 &  30.75 \\
 No Perc. Loss &  0.0122 & 20.57 &  0.83 &  33.65 \\
 No Latent Loss &  0.0125 &  20.51 &  0.84 &  30.86 \\
 No Style Loss &  0.0123 &  20.56 &  0.84 &  34.41 \\
 No Temporal &  0.0111 &  21.60 &  0.86 &  27.67 \\
 Full &  \textbf{0.0101} &  \textbf{22.81} &  \textbf{0.87} &  \textbf{26.92} \\

\hline
\end{tabular}
% \vspace{-2mm}
\end{center}
\caption{Ablation studies for different components on the coloring results, using  mean MSE, PSNR, SSIM and FID. }\label{table:acc-input}
% \vspace{-3mm}
\end{table}

\begin{table}
\begin{center}
\begin{tabular}{|l|c|c|c|c|}
\hline
Method & MSE$\downarrow$ & PSNR$\uparrow$ & SSIM$\uparrow$ & FID$\downarrow$ \\
\hline%\hline
cGAN~\cite{DBLP:journals/corr/HensmanA17} & 0.0366 & 15.07 & 0.72 & 63.48 \\
TCVC~\cite{thasarathan2019automatic} & 0.0426 & 14.95 & 0.73 & 50.75 \\
Two-stage~\cite{zhang2019two-stage} & 0.0352 & 14.91 & 0.65 & 42.08 \\
Deep IA~\cite{Liao:2017:VAT:3072959.3073683} & 0.0478 & 15.36 & 0.67 & 38.22 \\
Ours & \textbf{0.0104} & \textbf{22.41} & \textbf{0.86} & \textbf{27.93} \\
\hline
\end{tabular}
% \vspace{-2mm}
\end{center}
\caption{Quantitative comparison
with~\cite{DBLP:journals/corr/HensmanA17,Liao:2017:VAT:3072959.3073683,thasarathan2019automatic,zhang2019two-stage} using mean MSE, PSNR, SSIM and FID. Some examples of visual comparison are shown in Fig.~\ref{fig:ana-other}.}
\label{table:acc}
% \vspace{-5mm}
\end{table}

\subsection{Ablation Studies}

We perform ablation studies to evaluate the contribution of each module and loss term. The quantitative results comparing our full pipeline with one component disabled are reported in Table ~\ref{table:acc-input}, using standard metrics PSNR, MSE, SSIM, Fr\'{e}chet Inception Distance (FID)~\cite{heusel2017gans}. This shows that the similarity-based color transform layer, Embedder, usage of distance field maps, perceptual loss, latent loss, style loss, and filter are all essential to the performance of our method. 
\jqnew{When testing, we use the first and last frames of the video sequences as reference images to color all the intermediate frames.} 
The visual comparison of an example is shown in Figure~\ref{fig:ana-component}. 
\jqnew{Figure~\ref{fig:ana-component} shows that the $Sim$ module greatly contributes to the generation result. The $Emb$ module constrains the generation result to match the color style of the reference images (see the region of human hair). The temporal constraint network ensures that the colors of resulting images are more chronologically coherent. We conducted a further experiment to validate the importance of each part of the network, by coloring the target frame using features learned with only certain sub-modules. As shown in Figure~\ref{fig:color-decoder}, when the target image and the reference image have large deformation, just using features from $Sim$ is insufficient to generate accurate results, and adding $Emb$ preserves the global color style of reference images very well.

In experiments, we found that adding spectral normalization to the network can improve the stability of the training, but adding the spectral normalization to the Emb module of the color transform network will cause the generation results to be dim and the colors not bright. Figure~\ref{fig:color-sim} shows the results of Sim reconstruction and network generation in the following three cases. The color transform network all adds spectral normalization, none of them adds spectral normalization, and only the $Emb$ module does not add it. It can be seen from Figure~\ref{fig:color-sim} that the final result of adding the spectral normalization network to the $emb$ module is dim, and $Sim$ cannot learn specific color information without adding it.
}

\begin{table}
\begin{center}
\begin{tabular}{|l|c|c|c|c|}
\hline
Method & MSE$\downarrow$ & PSNR$\uparrow$ & SSIM$\uparrow$ & FID$\downarrow$ \\
\hline%\hline
Basic ~\cite{thasarathan2019automatic}- & 0.0523 & 13.32 & 0.63 & 78.64 \\
Fine-tuned ~\cite{thasarathan2019automatic}- & 0.0282 & 16.17 & 0.71 & 72.12 \\
Basic ours- & 0.0132 & 19.57 & 0.85 & 66.72 \\
Fine-tuned ours- & \textbf{0.0073} & \textbf{23.03} & \textbf{0.89} & \textbf{33.71} \\
\hline 
Basic ~\cite{thasarathan2019automatic}+ & 0.0951 & 10.71 & 0.56 & 86.04 \\
Fine-tuned ~\cite{thasarathan2019automatic}+ & 0.0584 & 12.94 & 0.59 & 77.81 \\
Basic ours+ & 0.0197 & 18.06 & 0.82 & 68.23 \\
Fine-tuned ours+ & \textbf{0.0138} & \textbf{20.50} & \textbf{0.86} & \textbf{37.79} \\
\hline

\end{tabular}
% \vspace{-3mm}
\end{center}
\caption{Quantitative evaluation with ~\cite{thasarathan2019automatic} using mean MSE, PSNR, SSIM and FID on animation Dragon Ball. ``-'' indicates a short sequence of length 8 is used for testing; ``+'' indicates the entire sequence is tested, regardless of the length of the sequence. }\label{table:acc-ball}
% \vspace{-3mm}
\end{table}
\subsection{Comparisons of Line Art Image Colorization}

We compare our approach to the state-of-the-art line art image coloring methods, cGan-based~\cite{DBLP:journals/corr/HensmanA17}, Deep Image Analogy (IA)~\cite{Liao:2017:VAT:3072959.3073683}, Two-stage method~\cite{zhang2019two-stage}, and TCVC~\cite{thasarathan2019automatic}. For fairness, we use the training set described in Sec. ~\ref{sec:DataCollection} for  all methods and then evaluate them on the same test set, and only one reference image is used for each sequence as other methods cannot take multiple references. 
\jqnew{In order to conduct a fair comparison, for the two reference images needed in our network, we both use the first frame of a video sequence.} Figure~\ref{fig:ana-other} shows the coloring results of all the methods. We further use several standard metrics to quantitatively evaluate the coloring results, which are presented in Table~\ref{table:acc}. The experimental results show that our method not only gets better coloring results, but also maintains the style consistency with the reference image. 

\begin{table}
\begin{center}
% \hspace{-2mm}
\begin{tabular}{|c|c|c|c|c|c|c|}
\hline
Anime & Method & MSE$\downarrow$ & PSNR$\uparrow$ & SSIM$\uparrow$ & FID$\downarrow$   \\
\hline%\hline
% \multicolumn{5}{|c|}{Amanchu} \\
% \hline
\multirow{2}{*}{3} & ~\cite{thasarathan2019automatic} & 0.0345 & 15.23 & 0.66 & 76.34 \\
& Ours & \textbf{0.0105} & \textbf{20.91} & \textbf{0.83} & \textbf{51.99} \\
\hline
% \multicolumn{5}{|c|}{SSSS.GRIDMAN} \\
% \hline
\multirow{2}{*}{4} & ~\cite{thasarathan2019automatic} & 0.0412 & 14.52 & 0.66 & 99.13 \\
&Ours & \textbf{0.0090} & \textbf{22.16} & \textbf{0.85} & \textbf{56.30} \\
\hline
% \multicolumn{5}{|c|}{NO.6} \\
% \hline
\multirow{2}{*}{5} & ~\cite{thasarathan2019automatic} & 0.0455 & 14.67 & 0.59 & 95.68 \\
&Ours & \textbf{0.0095} & \textbf{21.91} & \textbf{0.82} & \textbf{58.27} \\
\hline
% \multicolumn{6}{|c|}{Soul Eater} \\
% \hline
\multirow{2}{*}{6} & ~\cite{thasarathan2019automatic} & 0.0585 & 13.14 & 0.62 & 91.38 \\
&Ours & \textbf{0.0113} & \textbf{20.97} & \textbf{0.83} & \textbf{54.52} \\
\hline
\end{tabular}
\end{center}
% \vspace{-2mm}
\caption{Quantitative evaluation of different anime videos, using 32 video sequences to fine tune network parameters. }
\label{table:acc-anime}
% \vspace{-3mm}
\end{table}
\subsection{Comparisons on Video Colorization}

We compare our method with other colorization methods for animation coloring, as shown in 
Figure~\ref{fig:ana-ball}. We select 32 sequences from Anime 2 to fine-tune models of all methods and then test them on the remaining sequences. When coloring each sequence, the first image of the sequence is treated as the reference image, and the coloring results of the remaining images are evaluated. For the cGan-based~\cite{DBLP:journals/corr/HensmanA17}, Deep Image Analogy~\cite{Liao:2017:VAT:3072959.3073683}, and Two-stage generation~\cite{zhang2019two-stage} methods, the selected reference image is used to color the remaining images in the sequence. For TCVC~\cite{thasarathan2019automatic}, the selected  frame in a sequence is first used as the condition input to color the second frame, and the subsequent frames use the generation result of the previous frame as the condition input. It can be seen that TCVC~\cite{thasarathan2019automatic} keeps propagating color errors to subsequent frames, leading to error accumulation, while our method can effectively avoid this problem. The cGan-based~\cite{DBLP:journals/corr/HensmanA17} and Deep Image Analogy~\cite{Liao:2017:VAT:3072959.3073683} methods cannot handle the line art image coloring with large differences in shape and motion. The result of Two-stage generation~\cite{zhang2019two-stage}  does not maintain consistency with the reference images well.

Our model trained on one animation also generalizes well to new animations. When their color styles are different, our method benefits from using a small number of sequences from the new animation for fine tuning. To demonstrate this, we apply the network trained on Anime 1 to color Anime 2 (basic model), with optional fine-tuning of the network using 32 sequences from Anime 2 (fine-tuned model). 
We compare the results with  TCVC~\cite{thasarathan2019automatic}, using the same settings. As shown in Table~\ref{table:acc-ball}, our method 
achieves better coloring results after fine-tuning the network with only a small amount of new animation data. Table~\ref{table:acc-anime} shows the quantitative testing results  of our method and TCVC on other 4 anime videos where the models are fine-tuned by 32 video sequences. This demonstrates the strong generalizability of our network to unseen animation styles. 
Our method outperforms TCVC~\cite{thasarathan2019automatic} by a large margin in all settings. 
%%%zfl: delete this since it has been mentioned?
%\jqnew{For fairness testing, our method uses only the first frame as the reference image.}

\begin{figure}\centering
	{\includegraphics[width=0.98\linewidth]{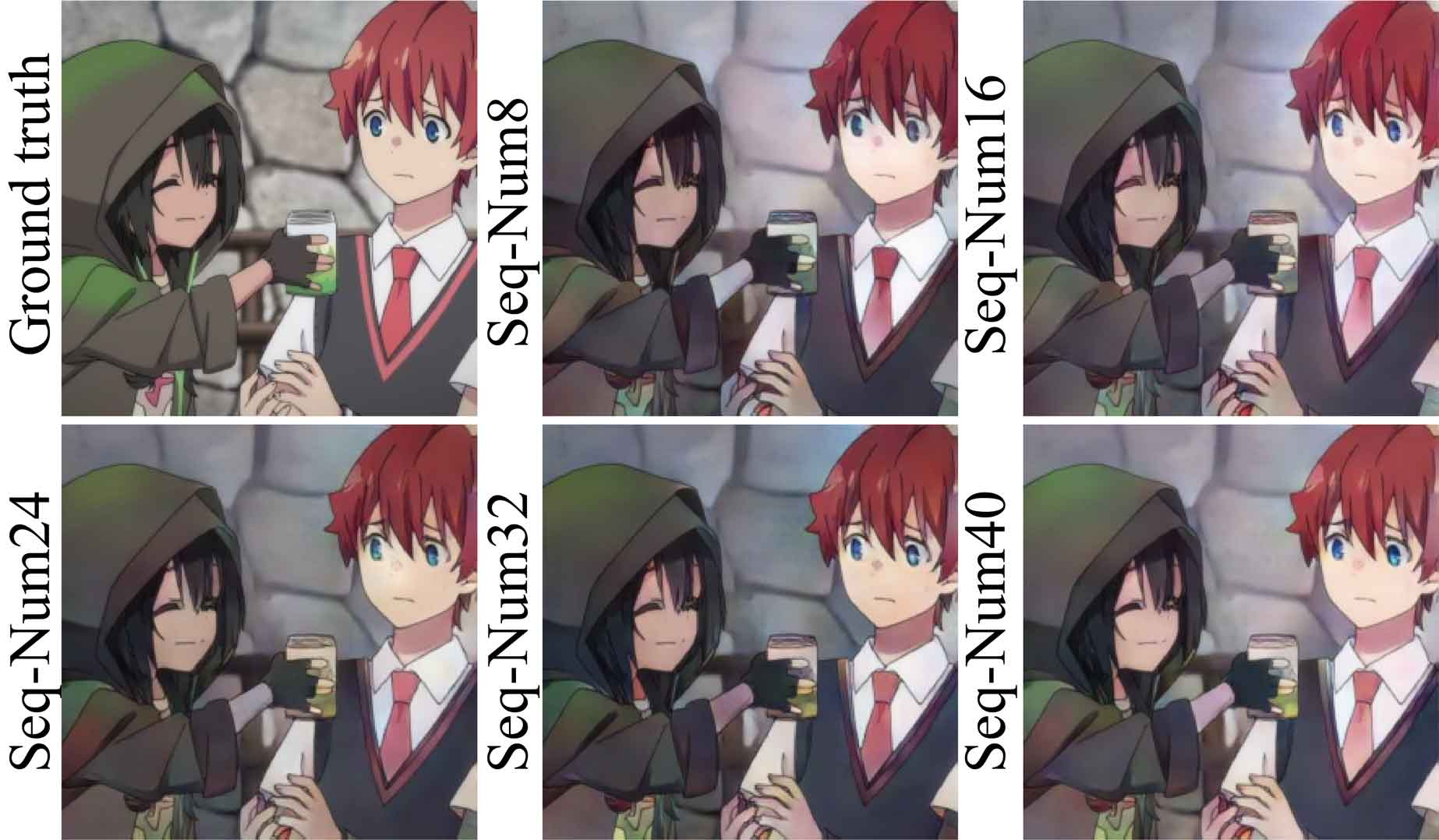}}

    \caption{Comparison of coloring results of the network after inputting different numbers of images to fine tune the network. }
	\label{fig:ana-finenums}
% 	\vspace{-3mm}
\end{figure}

\subsection{More Discussions}\label{sec:MoreDis}

\jqnew{
\textbf{Number of reference images.} We test the effect of feeding different numbers of reference images to train the model for coloring long sequences. We divide a video sequence into multiple segments according to different numbers of reference images. We use the frames where the sequence is divided as the reference images to color the in-between frames. We set the reference image number as $K = 1, 2, \dots, 5$ respectively. The coloring are shown in Figure~\ref{fig:ana-diffseq}, where the quality is improved with increasing number of reference images. The quantitative evaluation is shown in Table~\ref{table:acc-preref}. It can be seen that more reference images generally lead to better results, but when $K$ exceeds 3, the improvement of the coloring results starts to converge. To balance between the reference image numbers and the coloring quality, we choose $K$ as 3 when applying our method to long time sequence colorization.
}

\textbf{Number of sequences for fine tuning.} For new animations, if the style is similar to the training dataset, we can get high quality color results. But for anime with a different style, a small amount of data is needed to fine tune the network to learn new styles. We test different numbers of sequences (Seq-Num) to fine tune the network parameters to achieve satisfactory coloring results in a new style, where Seq-Num is set to 8, 16, 24, 32, 40 respectively. 
\zjq{The fine-tuning phase takes about 2 hours of training for 30 epochs.} As shown in Figure~\ref{fig:ana-finenums} and quantitative comparison in Table~\ref{table:acc-diffseg},
the network generally produces better coloring results with increasing Seq-Num, and the results stabilize when Seq-Num exceeds 32. 
Thus, to balance between the coloring results and required training data amount, we set Seq-Num to 32 by default.

\begin{table}
\begin{center}
\begin{tabular}{|l|c|c|c|c|}
\hline
$K$ & MSE$\downarrow$ & PSNR$\uparrow$ & SSIM$\uparrow$ & FID$\downarrow$\\
\hline%\hline
1 & 0.0170 & 19.87 & 0.83 & 31.09 \\
2 & 0.0111 & 21.59 & 0.86 & 27.67 \\
3 & 0.0084 & 22.76 & 0.88 & 24.18 \\
4 & 0.0069 & 23.44 & 0.89 & 22.20 \\
5 & \textbf{0.0062} & \textbf{23.83} & \textbf{0.89} & \textbf{21.79} \\
\hline
\end{tabular}
% \vspace{-2mm}
\end{center}
\caption{Quantitative comparison of different numbers of reference images $K$ on the coloration of video sequences. 
}\label{table:acc-preref}
% \vspace{-3mm}
\end{table}

\begin{table}
\begin{center}
\begin{tabular}{|l|c|c|c|c|}
\hline
Setting & MSE$\downarrow$ & PSNR$\uparrow$ & SSIM$\uparrow$ & FID$\downarrow$ \\
\hline
Seq-Num8 & 0.0204 & 18.19 & 0.78 & 85.81 \\
Seq-Num16 & 0.0157 & 19.44 & 0.80 & 78.42 \\
Seq-Num24 & 0.0147 & 19.91 & 0.81 & 77.12 \\
Seq-Num32 & \textbf{0.0127} & 20.21 & \textbf{0.83} & 71.84 \\
Seq-Num40 & 0.0128 & \textbf{20.55} & 0.83 & \textbf{70.74} \\
\hline
\end{tabular}
% \vspace{-2mm}
\end{center}
\caption{Quantitative evaluation of fine-tuning with different sequence numbers.}
\label{table:acc-diffseg}
% \vspace{-4mm}
\end{table}

\section{Conclusions}

In this paper, we propose a new line art image video coloring method based on  reference images. The architecture exploits both local similarity and global color styles for improved results. Our method does not use sequential propagation to color consecutive frames, which avoids error accumulation. We collect a dataset for evaluating line art colorization. Extensive evaluation shows that our method not only performs well for coloring on the same video used for training, but also generalizes well to new anime videos, and better results are obtained with fine-tuning with a small amount of data.

\bibliographystyle{unsrt}
\bibliography{ieebib}

% \begin{IEEEbiography}{Michael Shell}
% Biography text here.
% \end{IEEEbiography}

% % if you will not have a photo at all:
% \begin{IEEEbiographynophoto}{John Doe}
% Biography text here.
% \end{IEEEbiographynophoto}

% % insert where needed to balance the two columns on the last page with
% % biographies
% %\newpage

% \begin{IEEEbiographynophoto}{Jane Doe}
% Biography text here.
% \end{IEEEbiographynophoto}

% or
%\appendix  % for no appendix heading
% do not use \section anymore after \appendix, only \section*
% is possibly needed

% use appendices with more than one appendix
% then use \section to start each appendix
% you must declare a \section before using any
% \subsection or using \label (\appendices by itself
% starts a section numbered zero.)
%

% \appendices
% \section{Proof of the First Zonklar Equation}
% Appendix one text goes here.

% % you can choose not to have a title for an appendix
% % if you want by leaving the argument blank
% \section{}
% Appendix two text goes here.

% % use section* for acknowledgment
% \ifCLASSOPTIONcompsoc
%   % The Computer Society usually uses the plural form
%   \section*{Acknowledgments}
% \else
%   % regular IEEE prefers the singular form
%   \section*{Acknowledgment}
% \fi

% The authors would like to thank...

% Can use something like this to put references on a page
% by themselves when using endfloat and the captionsoff option.
\ifCLASSOPTIONcaptionsoff
  \newpage
\fi

\end{document}